# A Bioinspired Underwater Robot with a Latch-Mediated Soft Bistable Mechanism

Chongze Bi†, Wenjie Wu†, Zonghao Zuo, and Li Wen*

*Abstract*— **Underwater robotics has advanced significantly over recent decades; however, the development of miniaturized underwater robots remains limited by low energy densities of traditional power sources. Nature offers compelling solutions—organisms like mantis shrimps and fleas utilize latch-mediated spring actuation (LaMSA) systems that achieve rapid movements through a decoupled energy storage and release mechanism. Despite extensive studies of LaMSA, replicating such rapid, asymmetric actuation within simple, compact structures remains challenging. In this work, we introduce a bioinspired, soft bistable actuator with an integrated latch mechanism that enables asymmetric energy input and release using a single motor. Coupled with fin structures, this design facilitates efficient underwater propulsion and maneuverability. Experimental results demonstrate stable periodic flapping, precise steering, and a maximum thrust of 0.528 N, impulse of 0.147 N·s, and vertical displacement of 30 mm. By modulating fin angles, the robot achieves versatile motions, including vertical ascent, diagonal forward movement, and lateral translation. This study presents a novel, energy-efficient approach for controlling motion in compact underwater robots, paving the way for advanced biomimetic designs with potential applications in exploration, environmental monitoring, and inspection.**

## I. INTRODUCTION

With the increasing exploration and utilization of aquatic environments such as oceans, lakes, and rivers, an expanding number of underwater robots have been deployed for environmental monitoring and scientific investigation [1]. However, in tasks requiring delicate manipulation and minimal disturbance, such as in situ observation of small aquatic organisms or underwater data collection, conventional large-scale underwater robots are often unsuitable. Their considerable size and hydrodynamic impact may disturb fragile ecosystems or resuspend sediment, thereby compromising observational accuracy and operational effectiveness. Consequently, miniaturized underwater robots have attracted significant attention for such applications [2,3].

Despite their advantages, small-scale robots face critical challenges related to limited energy availability. As the physical size decreases, the reduction in onboard battery capacity inherently constrains locomotion capability and endurance [4]. To address this limitation, researchers have turned toward bio-inspired designs, which offer promising advantages such as high energy efficiency, low noise, and adaptability to complex environments [5]. Many small organisms, including mantis shrimps and fleas, exhibit remarkable power amplification despite their diminutive size. For instance, mantis shrimps can strike with the speed of approximately 100km/h within only 2 milliseconds [6]. Similarly, fleas can accelerate at 3200m/s², achieving jumps that reach up to 50 times their body length [7].

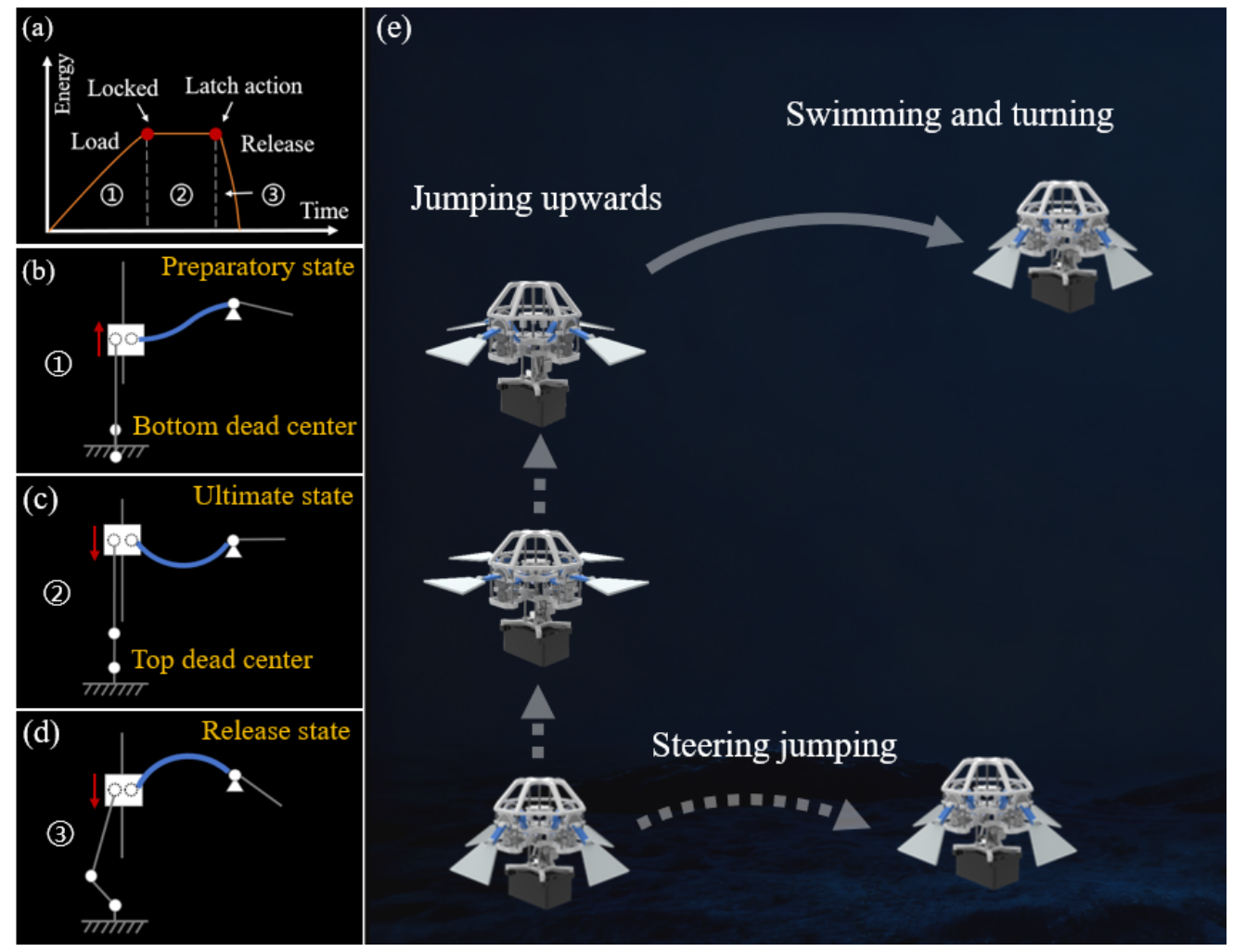

Figure 1. System description and Overview. a) Energy curve of a LaMSA system. b) Preparatory state of our robot. The sliding block is at bottom dead center. c) Ultimate state of our robot. The sliding block is at Top dead center. d) Releasa state of our robot. e) A Conceptual diagram of the overall robot locomotion process.

Extensive studies have revealed that latch mechanisms play a pivotal role in enabling such high-power movements. These biological systems can be classified as Latch-Mediated Spring Actuation (LaMSA) systems [8], characterized by a pronounced temporal asymmetry between energy loading and release. This configuration allows for low-power energy input combined with high-power output, thereby enabling small-scale organisms to achieve powerful and high-speed motions (Fig. 1 (a)).

The distinctive capabilities of LaMSA systems have inspired widespread research and their translation into miniature robotic platforms. Steinhardt et al. [9] investigated the rapid motion of mantis shrimps and developed a physical model elucidating how organisms and synthetic systems generate extremely high-acceleration, short-duration movements. Divi et al. [10] further explored how latch configurations regulate the conversion of stored potential energy into kinetic energy, examining how latch design and release mechanisms influence energy efficiency and environmental adaptability. They demonstrated these principles through a 4g jumper, validating the theoretical framework experimentally.

†These authors contribute equally to this work.
*Corresponding author: Li Wen (liwen@buaa.edu.cn)

Authors are with School of Mechanical Engineering and Automation, Beihang University

The power amplification effect inherent to LaMSA systems makes them particularly suitable for small-scale jumping robots. Zhao et al. [11] developed a lightweight multimodal robot weighing less than 30g, capable of both running and jumping to a height of 1.44 m. Francisco et al. [12] designed a centimeter-scale multimodal walking-jumping robot inspired by the flea's springtail. Their design, weighing 2.3g and measuring 6.1 cm in length, utilized a torque-reversal mechanism to rapidly release energy stored in a shape-memory-alloy actuator, achieving a jump distance of 23 body lengths. Hsiao et al. [13] integrated a LaMSA-based jumping leg with an active flapping-wing to create a microrobot capable of controlled takeoff, direction adjustment, and precision landing. Their design reduced overall energy consumption by 64% while increasing payload capacity by ten times.

Compared to continuous swimming, jumping locomotion exhibits remarkable terrain adaptability and obstacle negotiation capability [14]. As an emerging mode of motion, subaquatic jumping remains in the exploratory stage due to limitations such as low propulsion efficiency, high hydrodynamic resistance, and structural complexity [15].

Basing on the LaMSA system, this study proposes a flexible bistable actuator that serves as the elastic energy storage structure. The actuator incorporates a passive latch mechanism that enables slow, energy-efficient loading followed by rapid, high-power release using a single motor input. A miniature underwater robotic capable of jumping and swimming was then developed to experimentally validate the actuator's functionality.

## II. METHOD

The underwater robot primarily consists of a bistable actuator, a structural frame, four fins, and a sealed equipment compartment. The robot achieves movement through the coordinated flapping of four fins symmetrically distributed around the body, each driven by the bistable actuator.

The robot was designed with an emphasis on lightweight construction. The main structure is composed of 3D-printed components and carbon-fiber rods, resulting in a total weight of only 350 g. The bistable actuator is driven by a single DC motor, featuring a rotational speed of 167 RPM and a rated power of 4 W. Directional control and attitude adjustment are achieved through four miniature servos, each with a power output of 0.1 W.

### A. Design of Mechanism

Inspired by the latch mechanism of the mantis shrimp, which enables it to release tremendous energy within a fraction of a second to deliver a powerful strike, a latch-mediated bistable actuator was developed. The mantis shrimp utilizes a torque-reversal mechanism to rapidly release stored elastic energy, which is conceptually analogous to a bistable structure. Based on this biological inspiration, a hybrid design combining a latch mechanism with a bistable structure was proposed. Using this actuator to drive four fins, a compact underwater robot was designed, as shown in Fig. 2 (a).

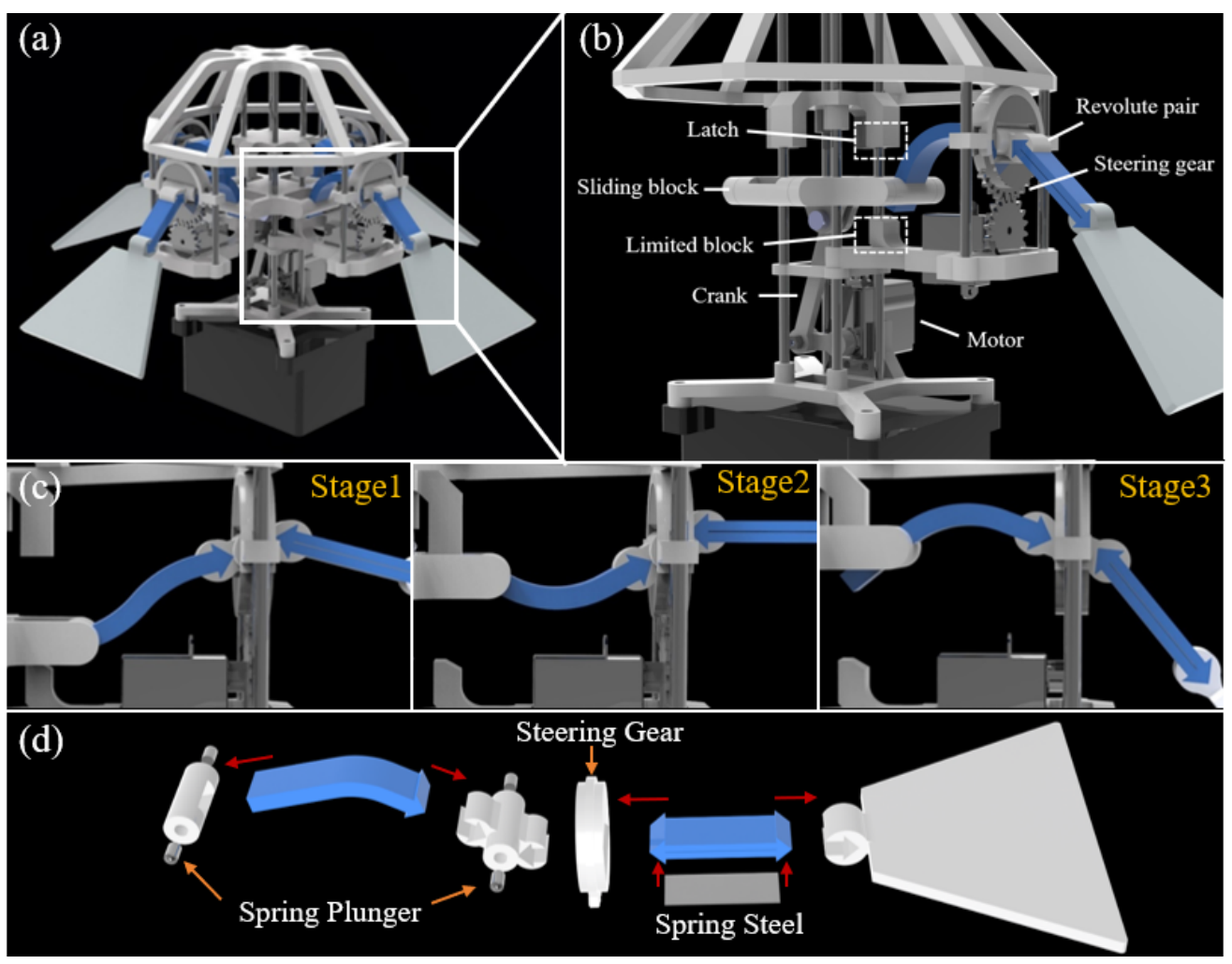

Figure 2. Mechanism and Structural Design. a) Overview of our robot. b) Mechanism design of the bistable actuator. c) Behavior of the bistable structure and fins at different stages. d) Assembly of main structures of the bistable actuator and the fin.

The core structure of the actuator is illustrated in Fig. 2 (b). A sliding block is constrained on a central carbon-fiber rod, allowing free vertical motion. On the edge of the sliding block, a flexible silicone rubber strip is connected via a revolute pair, while the other end of the strip is linked to a fin through another revolute pair. The sliding block is driven up and down by a motor through a crank–linkage mechanism, causing corresponding vertical motion of the silicone rubber strip. The silicone rubber strip serves as the bistable element of the system. Because its designed length exceeds the distance between its two joints, it becomes compressed and buckles when the sliding block moves upward. At this stage, the strip can be regarded as a pre-compressed bending beam exhibiting bistability. The bistable behavior of the silicone rubber strip is transmitted through the revolute pair and transformed into bistable changes in the flapping angle of the fins, thereby enabling rapid fin oscillation.

For the design of the fins, another silicone rubber strip is mounted between each fin and its revolute joint. Unlike the previous one, this strip embeds a thin spring-steel sheet to enhance stiffness. The purpose of this design is to suppress unwanted vibrations or hydrodynamic disturbances during motion, while ensuring efficient transmission of the force generated by the bistable actuator to the fins.

Moreover, owing to the intrinsic flexibility of silicone rubber, the bistable effect can still be maintained even under slight torsional deformation. Therefore, an additional revolute joint is incorporated at the fin connection, which is gear-coupled to a miniature servo. By adjusting the servo angle, the silicone rubber strip can be deflected slightly, generating a horizontal component of thrust. Through coordinated control of the four fins, the robot can achieve three degrees of freedom in attitude control, enabling directional movement, turning, and stable underwater maneuvering.

For a latch-mediated spring-actuation system, the essential characteristic lies in the temporal asymmetry between energy loading and release. In this design, a latch located near the top dead center (TDC) of the sliding block and a limited block at the bottom dead center (BDC) were incorporated to realize this asymmetric behavior. The actuation process can be divided into three stages, as shown in Fig. 2 (c). When the sliding block is at the BDC, the actuator is in the preparation stage, with the silicone rubber strip slightly compressed. The limited block constrains the strip to bend only downward, resulting in a downward fin angle. As the sliding block moves upward from the BDC to the TDC, the distance between the two revolute joints decreases, causing further compression of the strip and accumulation of elastic potential energy. When the sliding block approaches the TDC, the strip reaches its maximum compression, with the fin achieving its maximum angular displacement—the actuator is now at its extreme state.

As the sliding block continues to move upward, the strip comes into contact with the latch. The latch impedes further deformation and applies a downward reaction force on the strip. Once the sliding block surpasses the TDC, this reaction exceeds the threshold force required to trigger bistable switching. Consequently, the bistable structure snaps through, releasing the stored elastic energy in an extremely short time. The system is currently in release stage. During this snap-through, the fin angle switches abruptly, generating a reactive hydrodynamic force that propels the robot upward. After this, the sliding block moves from the TDC back to the BDC, and then begins the next cycle.

Throughout one actuation cycle, the motor's energy is gradually stored as elastic potential energy within the flexible silicone rubber strip and is then explosively released via bistable transition. The ratio of energy-loading time to energy-release time reaches approximately 10:1. This temporal asymmetry enables the fins to push the water downward rapidly during the power stroke and recover slowly during the return stroke, maximizing thrust while minimizing hydrodynamic drag during recovery.

## B. Design of Mechanism

According to the previous analysis, the flexibility of the bistable unit plays a decisive role in the actuator's performance. Fig. 2 (d) presents the assembly of main structures of the bistable actuator and the fin. A platinum-catalyzed addition-cure silicone rubber with a Shore hardness of 60 HA was selected as the base material for the bistable element. This material provides an optimal balance between elasticity and stiffness, ensuring that the generated force can be effectively transmitted to the fins while maintaining sufficient compliance for bistability.

The silicone rubber strip was fabricated through molding and curing. The pre-mixed silicone was poured into 3D-printed molds and cured at room temperature for 16 hours to form the desired geometry. To facilitate assembly and replacement, arrow-shaped mortise–tenon joints were designed at the end of the silicone rubber strip, allowing them to be easily inserted into the corresponding slots of the 3D-printed frame. For the fin-mounted silicone units, a thin spring-steel sheet was embedded during casting. Upon curing, the steel strip becomes tightly encapsulated within the silicone, reinforcing the stiffness of the structure.

The revolute pairs were designed to satisfy both lightweight and modularity requirements. To simplify assembly and ensure smooth rotational motion, two spring plungers were adopted as the revolute pairs.

# III. Experiment and Results

## A. Analysis of Bistable Structure

In this study, the bistable silicone element serves as an elastic energy storage component within a LaMSA system. It stores the energy supplied by a motor in the form of elastic potential energy and releases it rapidly through a bistable transition. The design and dimensional optimization of the bistable structure were systematically investigated to ensure efficient energy storage and release.

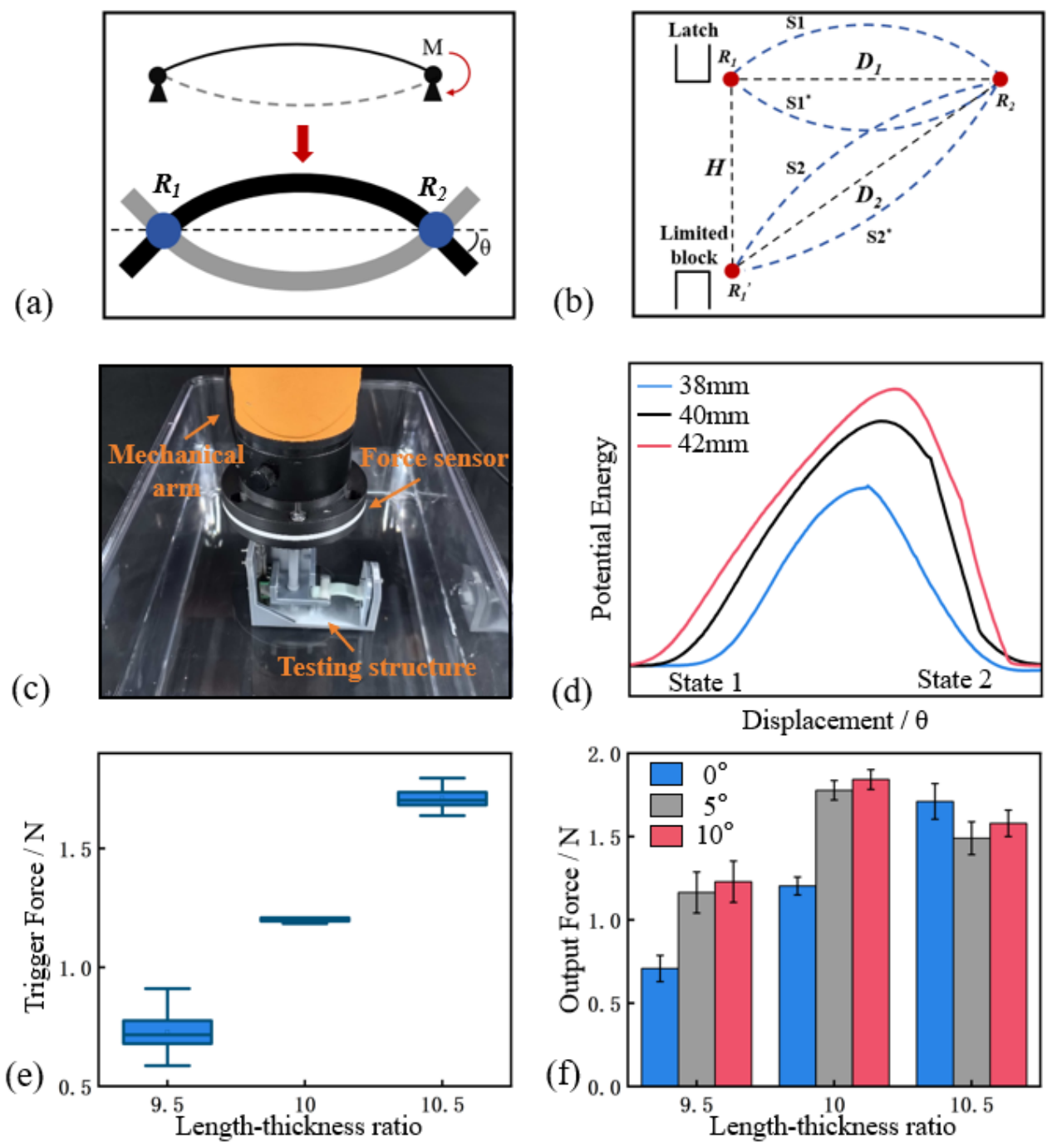

Figure 3. Modeling and Experimental Analysis of the Latch-mediated Bistable Structure. a) The differences between our bistable structure and the conventional pre-compressed bistable beam. b) A geometric model of bistable structure at different stages. c) Experimental setup. d) Potential energy curvy of the bistable structure. e) Comparasion of trigger force with different length-thickness ratios. f) Comparasion of output force with different length-thickness ratios and torsion angles.

Conventional pre-compressed bistable beams are typically fixed at both ends, which is effective only when the beam's length is much greater than its thickness. Moreover, such structures are often fabricated from rigid materials with relatively high stiffness, leading to limited deformation during state transition. In this study, flexible platinum-cured silicone rubber is employed as the beam material. Compared with rigid materials, silicone rubber offers greater deformability and improved fatigue resistance, making it well-suited for repeated bistable transitions. To accommodate large deformation, a revolute pair was introduced at both ends of the beam, as shown in Fig. 3 (a). The inclusion of revolute joints allows for angular bistability—during a state transition, both the curvature direction and the twisting angle of the beam's end change simultaneously.

For effective elastic energy storage, a high aspect ratio (length-to-thickness) is required. Experimental results revealed that this ratio has a crucial influence on bistable performance. In the experiments, the beam thickness was fixed at 4 mm, while its length was varied to determine the optimal configuration. The geometric schematic of the structure is shown in Fig. 3 (b), where $D_1$ denotes the distance between the rotational joints at the ultimate state, and $H$ represents the displacement range of the slider. When the system is at the BDC (i.e., the rotational joints at position $R_1'$), the silicone rubber beam is slightly compressed, exhibiting weak bistability between states $S_2$ and $S_2'$. The limited block allows the beam to remain at $S_2'$ with only a small holding force, thus we can calculate the lower limit of the length of the silicone rubber beam L.

$$L > D_2 \tag{1}$$

The maximum allowable beam length corresponds to the ultimate compressed state, where the joints are located at $R_1$ and the beam undergoes significant bending. In this case, the system exhibits strong bistability between states $S_1$ and $S_1'$. Based on previous works, to achieve a stable bistable effect, the pre-compression ratio should not exceed 30%.

$$0.7L < D_1 \tag{2}$$

Combining the geometric relationships yields the design constraint for the beam length.

$$\sqrt{D1^2 + H^2} < L < \frac{10}{7} D1 \tag{3}$$

Considering the extension section of the silicone rubber beam that interacts with the latch, the feasible range of beam length was determined to be 38mm to 42mm. Accordingly, three beam lengths—38mm, 40mm, and 42mm—were selected for testing, corresponding to length-thickness ratios of 9.5, 10, and 10.5, respectively. The experimental setup is shown in Fig. 3(c). A robotic arm was employed to drive the bistable beam, with a force sensor mounted at its end to record the variation in force during state transition. The potential energy characteristics of the flexible bistable beam (Fig. 3a) were evaluated by actuating one rotational joint and monitoring the force response. Since the deformation-induced restoring force is conservative, the measured force was used as a qualitative representation of the elastic potential energy. The resulting potential energy profiles are shown in Fig. 3(d), revealing two distinct energy minima (*State 1* and *State 2*), corresponding to the two stable states of the system. As the beam length increases, the energy barrier between the two states also increases, indicating that a larger input energy is required to trigger the transition, while the corresponding output energy also rises. This observation is consistent with the experimental results presented in Fig. 3(e) and Fig. 3(f). Specifically, both the trigger force and output force increase with length-thickness ratio. Additionally, the effect of small angular torsion on output force was investigated (Fig. 3f). For beams with length-thickness ratios of 9.5 and 10, a slight torsion resulted in higher output force, attributed to the enhanced deformation and additional stored elastic energy. However, for the 10.5-length-thickness beam, torsion weakened the bistable effect due to excessive deformation, thereby reducing output force. Moreover, large angular torsion was found to eliminate bistability altogether. Based on these results, the 10- length-thickness -ratio beam (40 mm length) exhibited the most balanced performance and was thus selected for subsequent underwater robot experiments.

*B. Robot Locomotion Behavior with Different Fin Size*

To evaluate the performance of the bistable actuator, a series of experiments were conducted to assess the robot's locomotion capability. The robot achieves motion through the flapping of its fins, driven by the bistable actuator. Therefore, the size of the fins plays a critical role in determining overall locomotion Behavior. In this study, the fins were designed in a trapezoidal shape, with size ranging from 1000 mm² to 4500mm² in increments of 500mm². Although the bistable actuator itself is triggered almost instantaneously, the silicone rubber connecter introduces a certain degree of compliance, resulting in a relatively smooth and complete flapping motion rather than an abrupt one. Consequently, the analysis focused on two key performance metrics: the maximum instantaneous output force, and the total impulse generated over a full flapping cycle.

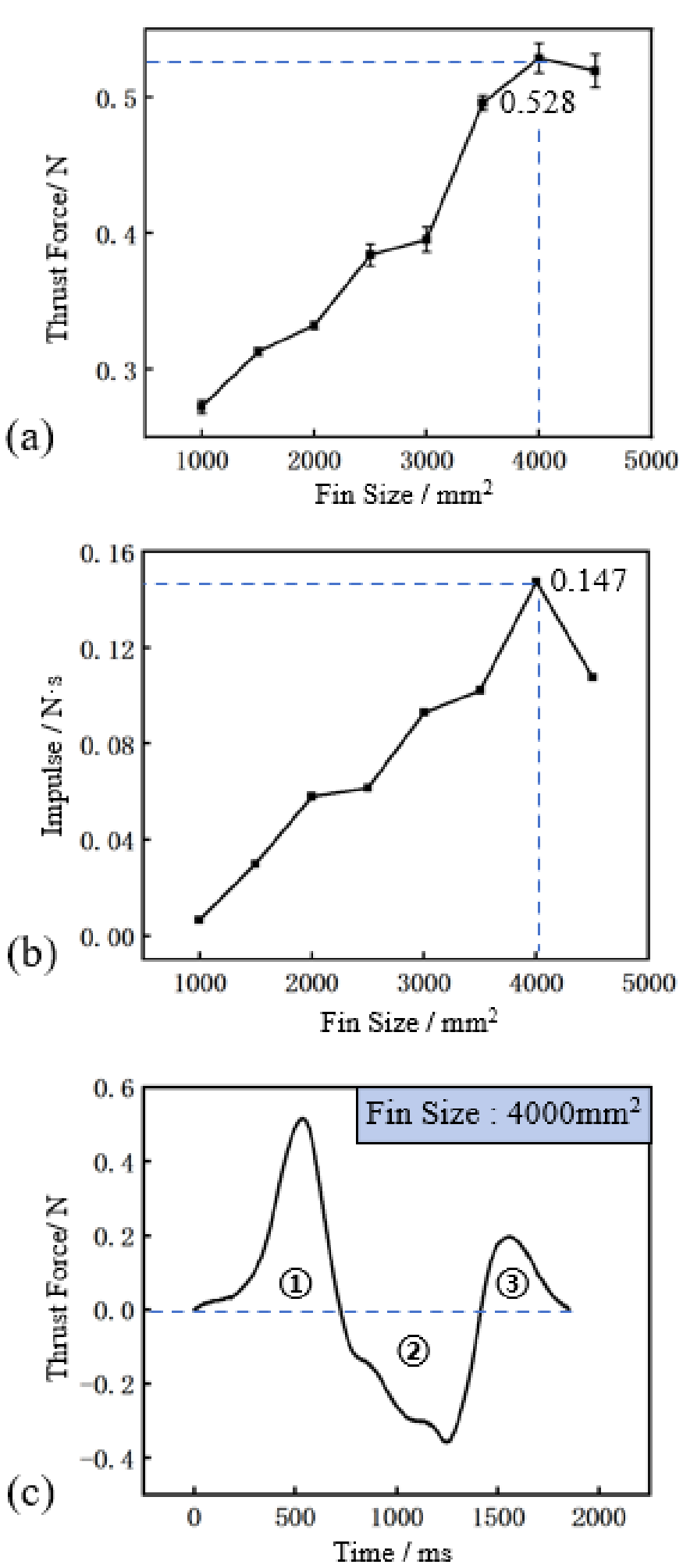

Figure 4. The influence of fin size on the locomotion behavior of robots. a) Comparasion of maximum thrust force with different fin sizes. b) Comparasion of impulse within a full locomotion cycle with different fin sizes. c) The variation of the thrust force over time within a full locomotion cycle.

The experimental results are presented in Fig. 4(a) and Fig. 4(b). As the fin size increased, both the thrust and impulse initially grew, reaching their peak values when the fin size was $4000 mm^2$, after which they began to decline. This trend suggests that larger fins can generate greater hydrodynamic force up to an optimal point, beyond which excessive surface area increases drag and inertia, reducing overall efficiency.

A detailed thrust force–time curve for one flapping cycle with a fin size of $4000 mm^2$ is shown in Fig. 4(c). Due to the inherent time asymmetry between the release and energy-loading phases of the bistable actuator, the curve exhibits a distinct asymmetric profile. During phase 1 (release phase), the fin rapidly moves downward, producing a sharp thrust peak of 0.528 N. During phases 2 and 3 (energy-loading phase), the fin gradually returns upward to its preparatory position. During this phase, the driving force initially remains directed downward, acting as resistance. As the fin approaches the preparatory position, inertial effects combined with the weak bistability caused by limited block cause a small upward force response.

## IV. Discussion

### A. Single Locomotion Analysis

Based on the previous results, the robot's locomotion performance was experimentally evaluated in a water tank. Fig. 5 illustrates the fin motion and the corresponding displacement of the robot over one actuation cycle, under both non-deflected and deflected fin configurations. Consistent with the analysis presented in Fig. 4(c), during each cycle the robot first moves upward as the fins flap downward during the power stroke. Subsequently, during the recovery stroke, the actuator exerts a downward force while gravity further decelerates the robot, resulting in a brief downward motion before the next cycle begins.

In the case of non-deflected fins, the robot achieved a maximum upward displacement of 40 mm, followed by a 10 mm downward motion from the peak, corresponding to a net periodic displacement of 30 mm per cycle. When the fins were deflected, the actuator generated a stronger output force, consistent with the experimental results shown in Fig. 3(f). Under this condition, the robot exhibited a maximum horizontal displacement of 27 mm and a maximum vertical displacement of 25 mm, demonstrating its capability for coupled translational motion.

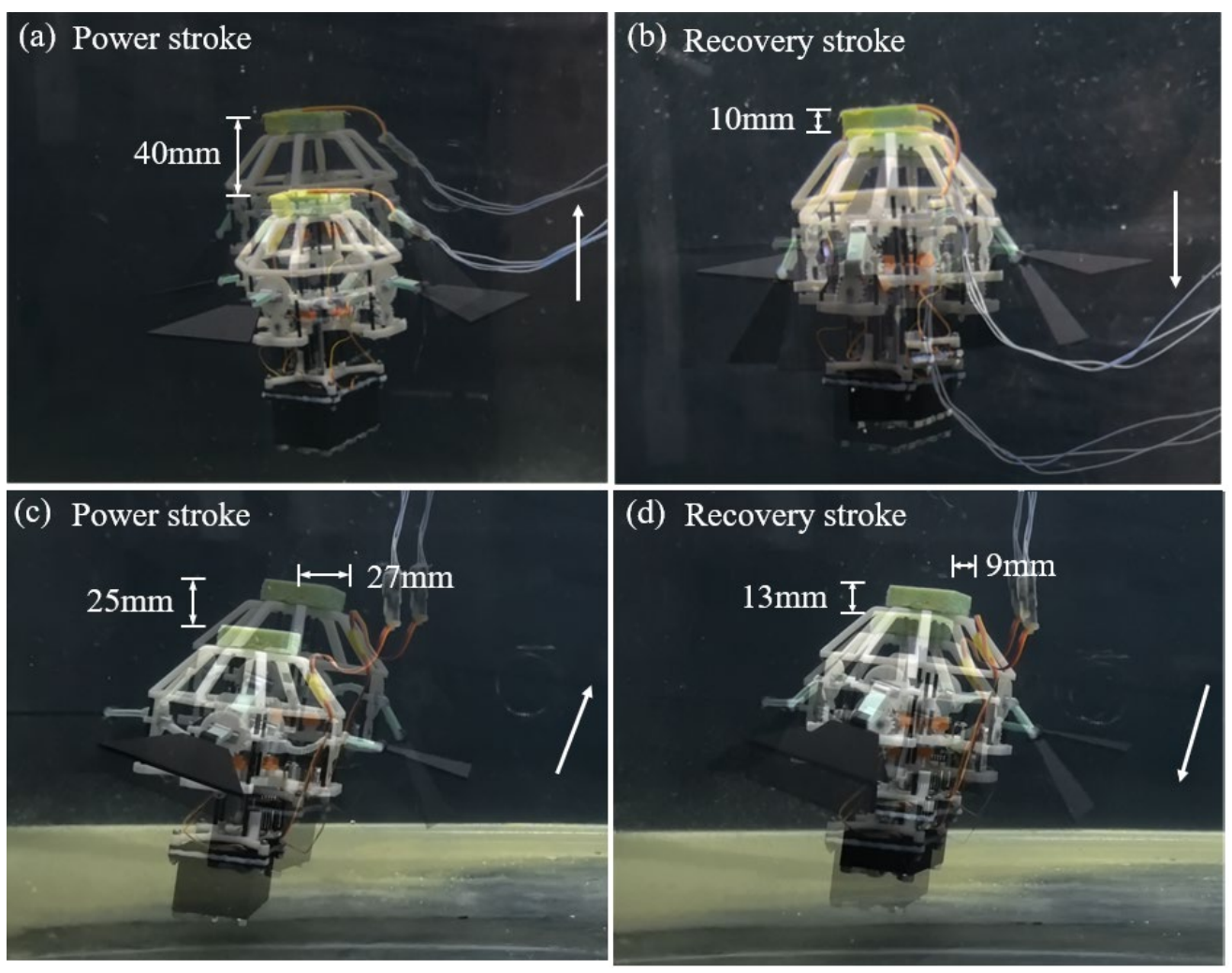

Figure 5. Locomotion schematic of the robot during a single cycle. a) Power stroke of the robot during the upward movement. b) Recovery stroke of the robot during the upward movement. c) Power stroke of the robot during the diagonal forward movement. d) Recovery stroke of the robot during the diagonal forward movement.

These results clearly reflect the design principles of the LaMSA system. Energy input occurs gradually during the recovery stroke, which occupies the majority of the cycle, while energy release is rapid and impulsive during the transition between bistable states. This inherent temporal asymmetry—slow energy loading and fast energy release—enables the bistable actuator to drive asymmetric periodic motion, thereby providing the robot with a high-power output capability within each actuation cycle.

### B. Steering Capability

As illustrated in Fig. 3(f), the flexible design of the bistable actuator enables slight torsional deformation of the fins without compromising thrust force. This feature not only preserves propulsion efficiency but also imparts the robot with attitude control and steering capability. Such functionality is achieved by employing four miniature servo motors, each connected through a steering gear to rotate the revolute joint at one end of the bistable structure.

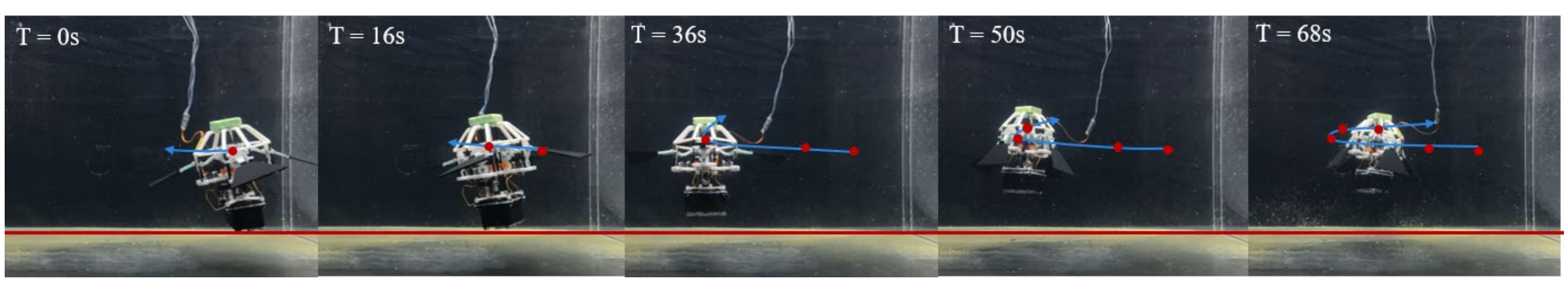

Figure 6. Steering capability of the robot. Robot swimming while turning.

Similar to a quadcopter's differential thrust mechanism, an increase in the horizontal component of the propulsion force inherently reduces the vertical thrust. Consequently, by adjusting the deflection angles of individual fins, the robot can perform various types of motion, including vertical ascent, diagonal forward motion, and lateral translation. Fig. 6 demonstrated the robot's steering capability. At the initial moment (T = 0s), the two lateral fins deflect symmetrically, altering the robot's pitch angle and inducing a leftward translation. Sixteen seconds later, differential fin deflection occurs, generating asymmetric thrust that causes the robot to turn right while moving forward. Subsequently, after completing the turn (T = 50s), the fin deflection angles are adjusted again, allowing the robot to move toward the right-forward direction.

The observed variations in both position and attitude confirm that the robot exhibits stable and controllable directional maneuvering within the water tank, validating the effectiveness of the bistable actuator's flexible design in enabling precise underwater orientation and trajectory control.

## V. Conclusion

In this paper, we introduced a novel design that integrates the latch-mediated spring actuation (LaMSA) mechanism, inspired by small organisms such as mantis shrimps, into an underwater robotic system. Based on the LaMSA concept, a soft bistable actuator was developed and implemented in a compact underwater robot. The actuator stores elastic potential energy through motor-driven compression of a flexible beam and releases it instantaneously via angular bistable switching. By integrating the soft bistable actuator with fin structures, the robot achieves slow energy input and rapid energy release in a compact and efficient configuration. The entire LaMSA-inspired actuation system is realized using a single motor, while four micro servos control fin rotation, enabling precise attitude and directional control of the robot.

The performance of the bistable actuator was experimentally characterized, and the optimal length-thickness ratio of the flexible silicone rubber beam was determined to be 10. The flexible beam design not only provides a large deformation range but also tolerates a certain degree of torsion. Experimental tests showed that the robot can perform stable periodic flapping and directional swimming in water. During each actuation cycle, the bistable actuator drives the fins to oscillate, generating a net forward thrust. The incorporation of controllable fin deflection further allows for pitch and yaw adjustments, enabling the robot to execute vertical ascent, diagonal forward motion, and lateral translation. The robot achieved a maximum thrust of 0.528 N, an impulse of 0.147 N·s, and a vertical displacement of 30 mm per cycle.

Overall, this study demonstrates the feasibility of integrating LaMSA systems into small-scale underwater robots. The proposed mechanism provides an effective strategy for achieving high-speed, high-efficiency, and controllable locomotion in compact underwater robots. Furthermore, it may influence future design and applications of miniaturized underwater robotic systems.